\documentclass[a4paper, 10pt, english]{article}
\usepackage[a4paper, total={6in, 9in}]{geometry}
\usepackage[export]{adjustbox}
\usepackage{amsmath}
\usepackage{amssymb}
\usepackage[english]{babel}
\usepackage{bm}
\usepackage{cancel}
\usepackage{caption}
\usepackage{color}
\usepackage{etoolbox}
\usepackage{float}
\usepackage{graphicx,graphics}
\usepackage{mathtools}
\usepackage{pgfplots}
\usepackage{pgfplotstable}
\usepackage{placeins}
\usepackage{subcaption}
\usepackage[skins]{tcolorbox}
\usepackage{tikz}
\usepackage{authblk}
\usepackage{orcidlink}
\pgfplotsset{compat=1.18}
\usepackage{hyperref}

\newcommand{\diff} [1]{\mathrm{d}{#1}}
\newcommand{\pdiff}[2]{\frac{\partial #1}{\partial #2}}
\newcommand{\n}{\bm{\nabla}}

\newcommand{\phia}{\phi_{\alpha}}

\renewcommand{\l}{\mathopen{}\mathclose\bgroup\left}
\newcommand{\e}{\aftergroup\egroup\right}

\let\originaleps=\epsilon
\let\epsilon=\varepsilon
\let\varepsilon=\originaleps

\title{evoxels: A differentiable physics framework for voxel-based microstructure simulations}

\date{}

\author[1]{Simon Daubner\orcidlink{0000-0002-7944-6026}\thanks{Corresponding author: \texttt{s.daubner@imperial.ac.uk}}}
\author[2]{Alexander E.~Cohen}
\author[3]{Benjamin Dörich\orcidlink{0000-0001-5840-2270}}
\author[1]{Samuel J.~Cooper\orcidlink{0000-0003-4055-6903}}

\affil[1]{Imperial College London, United Kingdom}
\affil[2]{Massachusetts Institute of Technology, United States}
\affil[3]{Karlsruhe Institute of Technology, Germany}

\begin{document}
\maketitle

\vspace{-1cm}
\section{Summary}
Materials science inherently spans disciplines: experimentalists use advanced microscopy to uncover micro- and nanoscale structure, while theorists and computational scientists develop models that link processing, structure, and properties. Bridging these domains is essential for inverse material design where you start from desired performance and work backwards to optimal microstructures and manufacturing routes. Integrating high-resolution imaging with predictive simulations and data‐driven optimization accelerates discovery and deepens understanding of process–structure–property relationships.

\begin{figure}[h!]
    \centering
    \includegraphics[width=0.9\linewidth]{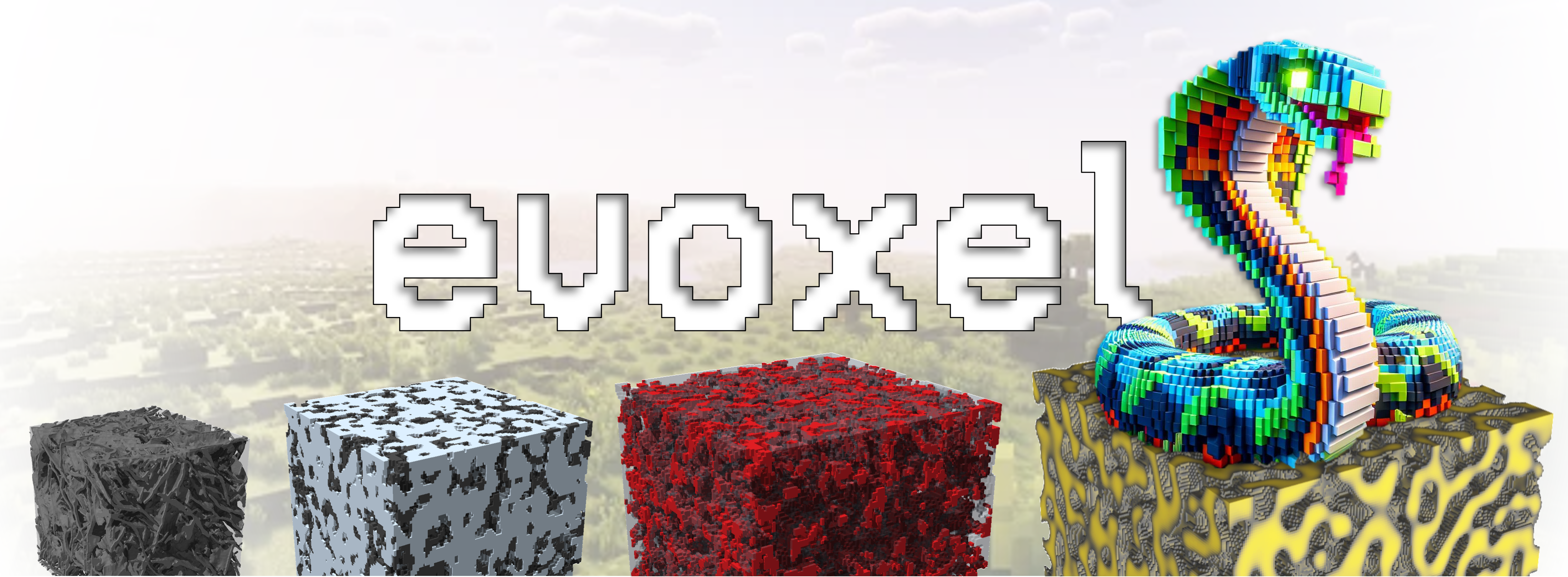}
    \caption{Artwork visualizing the core idea of evoxels: a Python-based differentiable physics framework for simulating and analyzing 3D voxelized microstructures.}
    \label{fig:graphical-abstract}
\end{figure}
The differentiable physics framework \textbf{evoxels} is based on a fully Pythonic, unified voxel-based approach that integrates segmented 3D microscopy data, physical simulations, inverse modeling, and machine learning.
\begin{itemize}
    \item At its core is a voxel grid representation compatible with both pytorch and jax to leverage massive parallelization on CPU, GPU and TPU for large microstructures.
    \item Both backends naturally provide high computational performance based on just-in-time compiled kernels and end-to-end gradient-based parameter learning through automatic differentiation.
    \item The solver design based on advanced Fourier spectral time-stepping and low-RAM in-place updates enables scaling to hundreds of millions of DOFs on commodity hardware (e.g. forward Cahn-Hilliard simulation with $400^3$ voxels on NVIDIA RTX 500 Ada Laptop GPU with 4GB memory) and billions of DOFs on high end data-center GPUs ($1024^3$ voxels on NVIDIA RTX A6000; more details see Figure~\ref{fig:benchmark}).
    \item Its modular design includes comprehensive convergence tests to ensure the right order of convergence and robustness for various combinations of boundary conditions, grid conventions and stencils during rapid prototyping of new PDEs.
\end{itemize}
While not intended to replace general finite-element or multi-physics platforms, it fills a unique niche for high-resolution voxel workflows, rapid prototyping for structure simulations and materials design, and fully open, reproducible pipelines that bridge imaging, modeling, and data-driven optimization.\\

\begin{figure}[h!]
    \centering
    \includegraphics[width=0.9\linewidth]{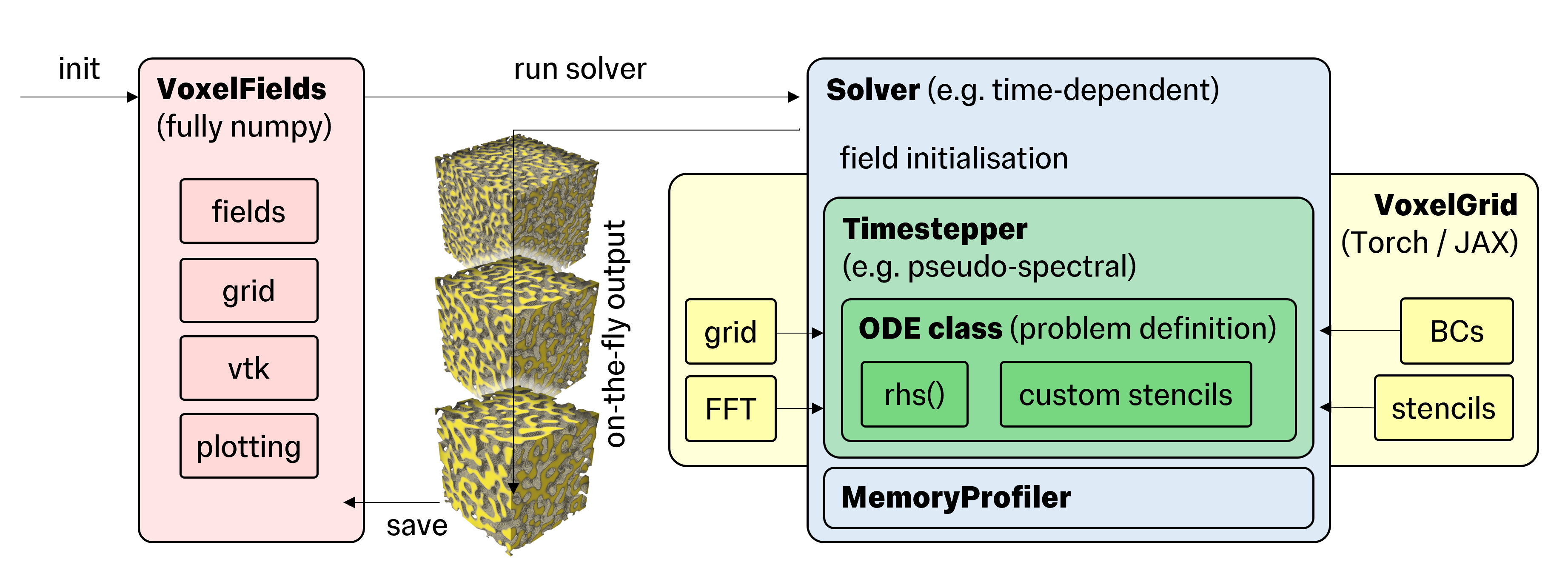}
    \caption{Visualisation of package concept. The VoxelFields class acts as the user interface for organising 3D fields on a regular grid including plotting and export functions. Solvers are assembled in a modular fashion. The chosen timestepper and ODE class are just-in-time compiled (green becomes one kernel) based on the given VoxelGrid backend.}
    \label{fig:code-schematic}
\end{figure}

From a high-level perspective, evoxels is organized around two core abstractions: VoxelFields and VoxelGrid. VoxelFields provides a uniform, NumPy-based container for any number of 3D fields on the same regular grid, maximizing interoperability with image I/O libraries (e.g. tifffile, h5py, napari, scikit-image) and visualization tools (PyVista, VTK). VoxelGrid couples these fields to either a PyTorch or JAX backend, offering pre-defined boundary conditions, finite difference stencils and FFT libraries as sketched in Figure~\ref{fig:code-schematic}. The implemented solvers leverage advanced Fourier spectral timesteppers (e.g. semi-implicit, exponential integrators), on-the-fly plotting and integrated wall-time and RAM profiling. A suite of predefined PDE “problems” (e.g. Cahn–Hilliard, reaction-diffusion, multi-phase evolution) can be solved out of the box or extended via user-defined ODEs classes with custom right-hand sides. Integrated convergence tests ensure each discretization achieves the expected order before it ever touches real microscopy data.

\textbf{evoxels} is aimed squarely at researchers who need a “plug-in-your-image, get-your-answer” workflow for digital materials science and inverse design. Experimentalists can feed segmented FIB-SEM or X-ray tomograms directly into high-performance simulations; computational scientists and modelers benefit from a truly open, reproducible framework. It speaks to anyone who wants special-purpose solvers for representative volume elements - without the overhead of mesh generation - while still offering the flexibility to develop new solvers, test boundary conditions, and incorporate machine-learning-driven optimization. evoxels provides both the turnkey usability of a specialized package and the extensibility of a low-level research toolkit for e.g. benchmarking tortuosity, fitting diffusion coefficients, or prototyping novel phase-field models.
\newpage

\section{Statement of Need}

Understanding the link between microstructure and material properties is a central challenge in materials science which increasingly relies on high-resolution 3D imaging, large-scale simulations, and data-driven optimization. Despite the growing availability of segmented volumes from FIB-SEM, X-ray CT, or synchrotron tomography, and data augmentation through generative AI~\cite{Kench2021,Finegan2022} the pipeline from experimental data to simulation remains fragmented. Existing simulation tools rarely operate directly on voxelized microscopy data, instead requiring costly meshing or complex preprocessing.
While boundary-conforming meshes (finite element/finite volume method) can better capture complex geometries, voxel-based methods (finite difference and Fourier pseudospectral methods)  -- especially in combination with smoothed boundary techniques~\cite{Yu2012,Daubner2024_micro} -- offer a robust and practical alternative for computing effective material properties.
In many materials science applications, small numerical or geometric errors (e.g. 5–10\,\%) are acceptable, as modeling assumptions are often approximate and the goal is to capture the correct order of magnitude or understand factors like tortuosity or relative transport rates -- that is, how much better or worse a given microstructure performs.
Furthermore, many commercial codes rely on proprietary data formats, complicating data exchange and reproducibility.
In addition to these technical hurdles, significant domain expertise is typically required to configure simulations i.e. choosing appropriate time-stepping schemes, numerical discretizations, and boundary conditions.
Even for well-studied problems such as the Cahn–Hilliard equation~\cite{Cahn1958, Zhu1999}, no scalable 3D Python implementation exists, highlighting a broader lack of open, reusable simulation frameworks in the field.
These gaps in data interoperability, code availability, and accessible expertise continue to hinder progress in understanding process-structure-property relationships and limit the practical deployment of inverse design methodologies.

The \textbf{evoxels} package enables large-scale forward and inverse simulations on uniform voxel grids, ensuring direct compatibility with microscopy data and harnessing GPU-optimized FFT and tensor operations.
This design supports forward modeling of transport and phase evolution phenomena, as well as backpropagation-based inverse problems such as parameter estimation and neural surrogate training - tasks which are still difficult to achieve with traditional FEM-based solvers.
This differentiable‐physics foundation makes it easy to embed voxel‐based solvers as neural‐network layers, train generative models for optimal microstructures, or jointly optimize processing and properties via gradient descent. By keeping each simulation step fast and fully backpropagatable, evoxels enables data‐driven materials discovery and high‐dimensional design‐space exploration.

There remains significant untapped potential in applying FFT-based semi-implicit schemes~\cite{Zhu1999} and exponential integrators~\cite{Hochbruck2010} across the broader landscape of digital materials science. Although these methods are well-established in areas such as spectral homogenization and phase-field modeling, their adoption has largely been limited to specialized research codes.
For example in \cite{CalCEOZ22}, a C\texttt{++}-CUDA implementation  of exponential integrators combined with FFT on a GPU was shown to outperform state-of-the-art exponential integrators implementations by fully exploiting the tensor structure of the spatial discretizations.
However, few open-source frameworks incorporate these methods into modern simulation pipelines that support automatic differentiation and GPU acceleration—capabilities increasingly critical for inverse design and data-driven workflows.

\begin{figure}[h]
\centering
\includegraphics[width=0.99\linewidth]{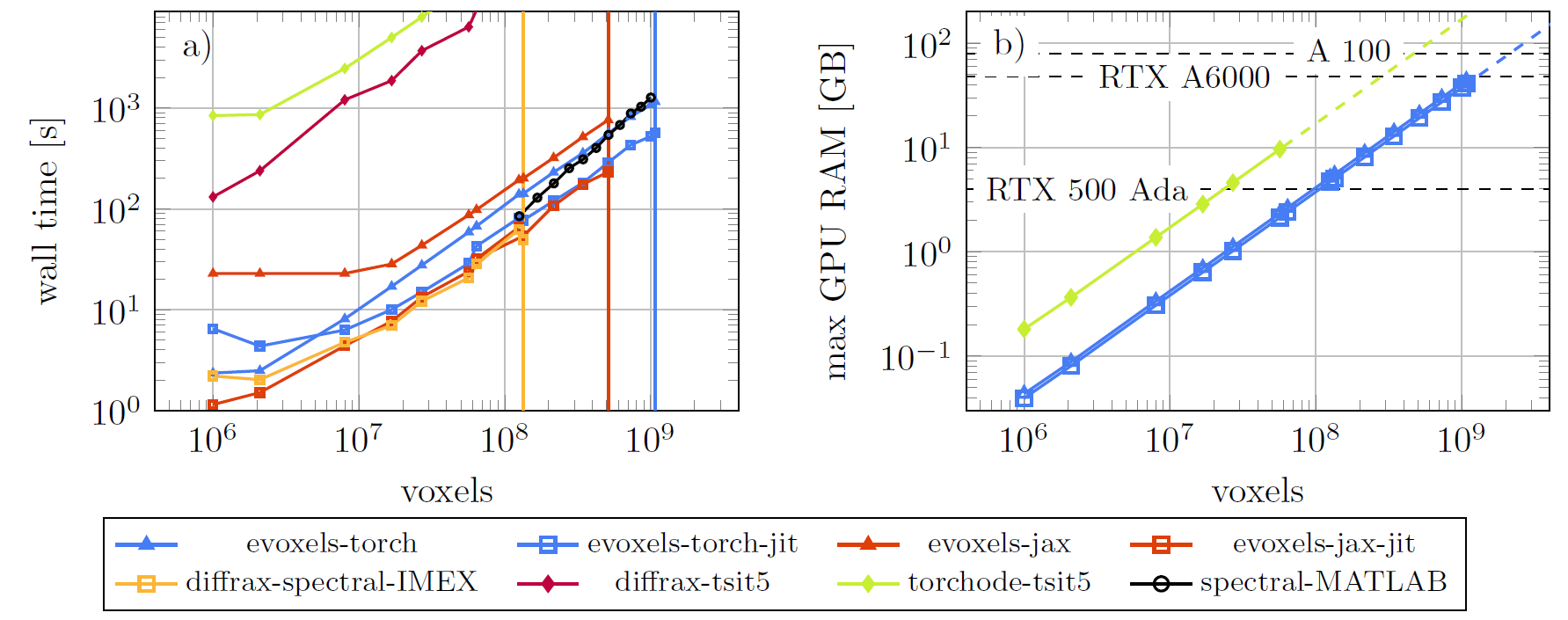}
\caption{Comparison of wall time and maximum GPU memory usage for the Cahn-Hilliard (CH) problem. Wall time for solving 1000 timesteps with fixed stepsize $\Delta t=1$ based on pseudo-spectral IMEX scheme with evoxels-torch (blue) and evoxels-jax (red) - both with and without just-in-time (jit) compilation; pseudo spectral IMEX scheme as custom diffrax solver (orange); and tsit5 scheme in combination with a PID timestep controller in torchode (green) and diffrax (purple). Vertical lines denote maximum problem size on Nvidia RTX A6000 for reference. Black datapoints refer to spectral element simulation of CH using MATLAB on Nvidia A100~\cite{XinyuLiu2024}. GPU memory footprint of all pytorch-based simulations shown in  b) shows linear scaling with amount of voxels.} \label{fig:benchmark}
\end{figure}

To evaluate performance against state-of-the-art python libraries, we benchmark the stiff, fourth‐order Cahn–Hilliard spinodal‐decomposition problem using torchode and Diffrax. As shown in Figure~\ref{fig:benchmark}, evoxels’ native pseudo‐spectral IMEX solver achieves runtimes one to two orders of magnitude shorter than general‐purpose ODE integrators and requires substantially less GPU memory. By contrast, the TSIT5 integrator with PID‐controlled timestepping which is available in both torchode and Diffrax demands finer timesteps, increasing both computation time and memory use to impractical levels for parameter optimization or inverse‐design tasks. We also provide a custom Diffrax pseudo‐spectral IMEX implementation fully integrated into the evoxels framework; while its wall time matches the native evoxels solver, it incurs higher memory overhead. Finally, fully implicit schemes (e.g., Diffrax’s Implicit Euler) exhaust GPU memory on moderate‐sized 3D grids (even $<100^3$), underlining their unsuitability for high-resolution microstructure simulations.

evoxels positions itself as a lightweight, accessible, and rigorously tested tool for prototyping voxel-based PDE solvers. Compared to domain-specific tools like taufactor~\cite{Kench2023} and magnum.np~\cite{Bruckner2023}, evoxels supports a broader range of problems, boundary conditions, and numerical methods while maintaining a modular, user-friendly interface for imaging-driven workflows.
At the same time, it is more specialized and efficient for problems on uniform grids with fixed physics than general-purpose solvers like FiPy or FEniCS.
evoxels is not intended to replace multiphysics platforms such as COMSOL or MOOSE, but to complement them by filling a niche in high-resolution, imaging-driven, and differentiable simulations.

Building on prior advances in microstructure characterization~\cite{Daubner2024_micro}, phase-field modeling for battery materials~\cite{Daubner2025} and the inverse learning of physics from image data~\cite{Zhao2023}, evoxels integrates these capabilities into a unified, extensible codebase. It is currently being used by researchers and students alike to advance inverse-learning capabilities and to develop advanced time integration methods.
In a field where open-source simulation tools remain underdeveloped, it provides a practical blueprint for reproducible digital materials science, helping to democratize capabilities that have long been confined to specialist groups or proprietary codebases.

\section*{Acknowledgements}

We acknowledge computational resources and support provided by the Imperial College Research Computing Service (http://doi.org/10.14469/hpc/2232). This work has received financial support from the European Union’s Horizon Europe research and innovation programme under grant agreement No 101069726 (SEATBELT project). Views and opinions expressed are however those of the author(s) only and do not necessarily reflect those of the European Union or CINEA. Neither the European Union nor the granting authority can be held responsible for them.
\newpage

\section*{Appendix: Microstructure modelling}
This section provides an insight into the PDEs behind the scenes, i.e. the types of computational problems that are commonly encountered in digital materials science and materials design. These problems can largely be classified into two categories:
one set governs the calculation of effective properties in complex heterogeneous structures; the other describes microstructural evolution via phase transformations, or reaction–diffusion mechanisms. In inverse‐design scenarios - where one prescribes target properties and then tailors the material’s developmental pathways - these two PDE classes become tightly coupled. Understanding and efficiently solving both types is essential for predictive modeling and optimized materials innovation.

\subsection*{Reaction-diffusion systems}
The time-dependent evolution of a single species undergoing reaction and diffusion given by Fick's law is described by the PDE
\begin{align}
\pdiff{c}{t}=\n\cdot\l(D(x,c) \n c \e) + f(c,t)
    \label{eq:reaction-diffusion}
\end{align}
where the diffusivity $D$ can be a function of space $x$ and/or composition $c$. Within the more general framework of linear irreversible thermodynamics, the driving force for diffusion is given by the gradient in chemical potential $\nabla\mu$ and with the chemical mobility $M$, the concentration evolution is then governed by
\begin{align}
\pdiff{c}{t}=\n\cdot\l(M(x,c) \n \mu(c) \e) + f(c,t).
    \label{eq:reaction-diffusion-general}
\end{align}
Note that the chemical potential is a function of concentration as in the case of the Cahn-Hilliard problem.

The Gray-Scott model is an example of two species coupled by an interaction term which can lead to surprisingly complex pattern formation.
It can be expressed as two coupled PDEs
\begin{align}
\pdiff{c_A}{t}&=\n\cdot\l(D_A\n c_A\e) - c_A c_B^2 + f(1-c_A),\\
\pdiff{c_B}{t}&=\n\cdot\l(D_B\n c_B\e) + c_A c_B^2 - k c_B.
\label{eq:gray-scott}
\end{align}
The model describes two species $A$ and $B$ which are dispersed in another medium assuming dilute solution (pairwise diffusion coefficients are equal to zero). $c_A$ is added at a given feed rate $f$ but it can maximally reach a concentration of $c_A=1$. A reaction converts $c_A$ into $c_B$ in the presence of $c_B$ (i.e. reaction term $c_A c_B^2$). $c_B$ is continuously removed with a given kill rate $k$ until it reaches $c_B=0$. Both species are also diffusing within the solution with given chemical diffusivities $D_A$ and $D_B$ respectively.

\subsection*{Phase transformations of first and second order}
We start from a free energy functional for a two-phase system
\begin{align*}
F_\text{int}=\int_{V} f_\text{int} \diff{V} = \gamma_0\int_{V}\epsilon|\n\phi|^2+\frac{9}{\epsilon}\phi^2(1-\phi)^2 \diff{V}
\end{align*}
where $\gamma_0$\,[J/m$^2$] denotes the interfacial energy and $\epsilon$\,[m] scales the width of the diffuse interface. Note that in this case, the homogeneous free energy is given by a double-well potential while in the context of the Cahn-Hilliard equation a regular free energy involving logarithmic terms is often employed and, in the context of the multiphase field method, an obstacle-type potential can be found. However, the following procedure would be identical in all cases.

The second order phase transition problem - also known as the \textit{Cahn-Hilliard equation}~\cite{Cahn1958,Cahn1961} - can be derived by inserting the chemical potential $\mu=\delta f_\text{int}/\delta\phi$ given by the functional derivative of the given free energy formulation into the mass conservation equation which yields
\begin{align}
    \pdiff{\phi}{t}=\n\cdot\l(M\n\mu\e) + f(\phi,t), \quad \mu= \gamma_0 g(\phi) - 2\gamma_0\epsilon\n^2 \phi
    \label{eq:cahn-hilliard}
\end{align}
which is a fourth-order PDE in space. 
Specifically, we use a variable mobility of the form $M=D_0 \phi(1-\phi)$ and the homogenous part of the chemical potential is given by the polynomial expression $g(\phi)=\frac{18}{\epsilon}\phi(1-\phi)(1-2\phi)$. Additionally, a forcing (source/sink) term $f$ can be considered.

Alternatively, the kinetics of a first-order phase transformation can be derived from the linear relaxation of the system free energy towards its minimum
\begin{align}
    \pdiff{\phi}{t}=-L\, \frac{\delta f_\text{int}}{\delta\phi}= \frac{M}{\epsilon} \l(2\gamma_0\epsilon\n^2 \phi - \gamma_0 g(\phi)\e)
    \label{eq:allen-cahn}
\end{align}
also known as the \textit{Allen-Cahn equation} for the non-conserved order parameter $\phi$. Note that the kinetic coefficient $L$ is chosen as $M/\epsilon$ such that $M$ represents a physical mobility of the migrating interface which is independent of its diffuse width scaled by $\epsilon$~\cite{Nestler2005}. Eq.~\eqref{eq:allen-cahn} leads to the formation of a diffuse interface and an evolution of the microstructure driven by curvature minimization.

An interesting variation of Eq.~\eqref{eq:allen-cahn} is obtained by subtracting the curvature-driven forces from the laplacian of $\phi$~\cite{Sun2007,Takaki2017,Schoof2020} such that
\begin{align*}
\pdiff{\phi}{t}=M\gamma_0\l(2\l(\n^2\phi-|\n\phi|\n\cdot\frac{\n\phi}{|\n\phi|}\e) - \frac{1}{\epsilon}g(\phi)\e).
\label{eq:allen-cahn-no-curvature}
\end{align*}
This will create a smooth transition in the direction of the surface normal while the shape is not altered by curvature minimization~\cite{Sun2007}.
Instead of subtracting the curvature the normal part of the laplacian can be computed directly~\cite{Schoof2020}
\begin{align*}
   \n^2\phi-|\n\phi|\n\cdot\frac{\n\phi}{|\n\phi|}= \n(\n\phi\cdot\bm{n})\cdot\bm{n}=\n|\n\phi|\cdot\frac{\n\phi}{|\n\phi|}.
\end{align*}

A multi-phase generalization of Eq.~\eqref{eq:allen-cahn} is given by a set of N coupled PDEs~\cite{Daubner2023,Hoffrogge2025} given as the sum of pairwise interactions
\begin{equation}
\pdiff{\phia}{t}=-\sum^{\tilde{N}}_{\beta\neq\alpha} \frac{M_{\alpha\beta}}{\epsilon\tilde{N}}\l(\frac{\delta f_\text{int}}{\delta\phi_\alpha} - \frac{\delta f_\text{int}}{\delta\phi_\beta} \e) \label{eq:multi_phase_evolution}
\end{equation}
where $\tilde{N}$ denotes the amount of locally present phases and $M_{\alpha\beta}$ denotes a matrix of pairwise interfacial mobilities.

\subsection*{Smoothed boundary method}
Oftentimes, the equations discussed above are confined to a given microstructure like lithium ion transport in the pore space of an electrode microstructure.
Following~\cite{Yu2012, Li2009}, we re-write the diffusion equation~\eqref{eq:reaction-diffusion} with the indicator function $\psi$ as
\begin{align}
    \psi\pdiff{c}{t}=\n\cdot\l(\psi D \n c \e) + |\n\psi| j_N + \psi f(c,t)
    \label{eq:c-smooth-boundary}
\end{align}
where $j_N$ is the normal boundary flux. The flux $j_N$ can vary spatially and/or temporally and for a closed system $j_N=0$ holds.
If the microstructure does not evolve over time ($\partial\psi/\partial t=0$) the equality $\psi\partial c/\partial t=\partial \psi c/\partial t$ holds. Therefore, we can re-formulate the PDE using $\psi c=z$
\begin{align}
    \pdiff{z}{t}&=\n\cdot\l(D_0 \n z - D_0 \frac{z}{\psi}\n \psi\e) + |\n\psi| j_N +\psi f(c,t)
    \label{eq:c-smooth-boundary-FFT}
\end{align}
This formulation is beneficial in terms of generality and the use of FFT based semi-implicit timestepping as discussed in the next section.

\subsection*{Semi-implicit timestepping}
Consider the (mass / heat) conservation equation (see Eq.~\eqref{eq:reaction-diffusion})
\begin{align*}
    \pdiff{u}{t}=\nabla\cdot\l(\Gamma(x,u)\nabla u\e) + f(u,t)
\end{align*}
with variable mobility (diffusivity/conductivity) $\Gamma$ and a forcing (sink/source) term $f$. Following the procedure sketched in~\cite{Zhu1999,Chen1998} we can re-formulate the equation to
\begin{align*}
    \pdiff{u}{t}=\nabla\cdot\l((\Gamma_0 + \Gamma(x,u) - \Gamma_0) \nabla u\e) + f(u,t)
\end{align*}
to then apply the semi-implicit Fourier spectral method to discretize the PDE in Fourier space
\begin{align*}
    \frac{\hat{u}^{n+1}+\hat{u}^n}{\Delta t} &= -k^2\Gamma_0 \hat{u}^{n+1} +k^2\Gamma_0 \hat{u}^{n} +\mathcal{F}\l\{\nabla\cdot\l(\Gamma(x,u)\nabla u\e) + f(u,t)\e\}^n
\end{align*}
where $\mathcal{F}(u)=\hat{u}$ denotes the Fourier transform.
This yields the following equation for the new timestep $\hat{u}^{n+1}$
\begin{align*}
    \hat{u}^{n+1}(1+\Delta t k^2\Gamma_0) &= \hat{u}^{n}(1+\Delta t k^2\Gamma_0) +\Delta t \mathcal{F}\l\{\nabla\cdot\l(\Gamma(x,u)\nabla u\e) + f(u,t)\e\}^n
\end{align*}
which corresponds to the following time-stepping scheme in real space
\begin{align*}
    u^{n+1} = u^n + \mathcal{F}^{-1}\l\{ \frac{\Delta t}{1+\Delta t k^2\Gamma_0} \mathcal{F}\l\{\nabla\cdot\l(\Gamma(x,u)\nabla u\e) + f(u,t)\e\} \e\}.
\end{align*}
Note that this procedure is relatively simple as it only involves coding the finite difference approximation of the original right-hand side of the problem $\text{rhs}=\nabla\cdot\l(\Gamma\nabla u\e) + f(u,t)$ as would be common to use any predefined timestepping scheme from existing python packages such as torchode and diffrax. In comparison to the explicit Euler scheme, this procedure additionally involves one forward FFT and one inverse FFT but leads to much larger stable timesteps~\cite{Zhu1999, Chen1998}. Note that as $u^{n}$ and $u^{n+1}$ both must satisfy the boundary conditions, the update $\mathcal{F}^{-1}\{\dots\}$ itself fulfills certain boundary constraints. Critically, as long as Dirichlet boundary conditions on $u$ are not a function of time the update exhibits constant zero boundary conditions which enables the efficient use of FFT.

This procedure can be equally applied to all reaction-diffusion and phase evolution problems given above, e.g. when applied to the fourth order Cahn-Hilliard problem, this results in
\begin{align*}
    \phi^{n+1} = \phi^n + \mathcal{F}^{-1}\l\{ \frac{\Delta t}{1+2\Delta t \epsilon M_0 k^4} \mathcal{F}\l\{\nabla\cdot\l(M\nabla \mu\e) + f(\phi,t)\e\} \e\}
\end{align*}
and, similarly, the smoothed boundary formulation of the diffusion equation then reads
\begin{align*}
    z^{n+1} = z^n + \mathcal{F}^{-1}\l\{ \frac{\Delta t}{1+\Delta t k^2 D_0} \mathcal{F}\l\{\nabla\cdot\l(D\nabla z -D\frac{z}{\psi}\nabla\psi\e) + |\nabla\psi| j_\text{N} + f(z,t)\e\} \e\}.
\end{align*}
The prefactor inside the inverse Fourier transform always takes the form $\Delta t / (1 - \Delta t\,\mathcal{S})$ where $\mathcal{S}$ is the symbol of the spatial operator i.e. its representation in the Fourier (spectral) domain. For instance, the diffusion operator $D\nabla^2$ corresponds to $-k^2D$.
Within the evoxels framework, the symbol must be defined together with the numerical right-hand side of the PDE for applying pseudo-spectral timesteppers to a given problem definition.

\end{document}